# Hand bone age estimation using divide and conquer strategy and lightweight convolutional neural networks

Amin Ahmadi Kasani[1], Hedieh Sajedi[2*]

[1]*Department of Mathematics, Statistics and Computer Science, College of Science, University of Tehran,* Tehran, Iran, aakasani@ut.ac.ir

[2]*Department of Mathematics, Statistics and Computer Science, College of Science, University of Tehran,* Tehran, Iran, hhsajedi@ut.ac.ir, Corresponding author

**Abstract** Estimating the Bone Age of children is very important for diagnosing growth defects, and related diseases, and estimating the final height that children reach after maturity. For this reason, it is widely used in different countries. Traditional methods for estimating bone age are performed by comparing atlas images and radiographic images of the left hand, which is time-consuming and error-prone. To estimate bone age using deep neural network models, a lot of research has been done, our effort has been to improve the accuracy and speed of this process by using the introduced approach. After creating and analyzing our initial model, we focused on preprocessing and made the inputs smaller, and increased their quality. we selected small regions of hand radiographs and estimated the age of the bone only according to these regions. by doing this we improved bone age estimation accuracy even further than what was achieved in related works, without increasing the required computational resource. We reached a Mean Absolute Error (MAE) of 3.90 months in the range of 0-20 years and an MAE of 3.84 months in the range of 1-18 years on the RSNA test set.

Keywords: Skeletal maturity, Deep neural networks, Machine learning, Bone Age Assessment, Computer aided diagnosis, Object detection.

## 1. INTRODUCTION

Measurement of bone maturity involves examining the size, shape, and degree of bone mineralization to find the distance between the current state of bone growth and full growth. Longitudinal growth in bones occurs through the process of endochondral ossification, whereas bone width increases with the development of skeletal tissue directly from the fibrous membrane. Calcification (the process of calcium deposition to produce bone) first begins near the center of the long bones in the primary ossification center. Many of the body's flat bones, including the wrist bone, are completely ossified from the primary ossification center. All long bones grow from the center of the secondary ossification, appearing in the cartilage at the end of the long bones like finger joints. In the process of children's growth, by assessing the areas of secondary and primary ossification center, the condition of bone growth and current growth can be assessed. Generally, children with normal growth conditions and of the same sex, with the same birth certificate age, have the same bone age [1][2].

Bone age assessment is a test for children that shows the difference between bone age and birth certificate age. Due to its practicality and high reliability, bone age assessment is a common and important approach for early detection of genetic disorders, several cases of pediatric developmental disorders, syndromes, and endocrine disorders, and response to disease treatment also provides physicians with information on children grows [3]. Research has shown that bone age can vary depending on race and place of residence, although this amount is not very high. In an Australian study, for example, using the Greulich-Pyle(GP) [4] method, the average bone age was 1.5 months for males and 3.7 months for females, less than their Chronological Age respectively [5].

Bone growth has several stages, the first stage is primarily in the first ossification center and usually ends in infancy, only a small proportion of them persist until the age of 10-year-old. The second stage, which lasts until the age of 18, mostly happens in the secondary ossification center. Most of the information obtained in estimating bone age is obtained from the analysis of changes in the ossification center. The process of bone production can be detected and examined by doctors and radiologists from left-hand radiographs. Diagnosis of children's bone age is of particular importance, used to diagnose genetic disorders, used to estimate a child's final height after puberty, due to the relationship between quality of life and final height, in many countries to assess the quality of life of Students are used in schools and gives doctors clues about endocrine disorders [2].

The methods used in the past, such as the GP method, require comparison and review of radiographic images in the method's atlas, by a radiologist, and the results for experienced radiologists are time-consuming and have low accuracy, similarly, The Tanner-Whitehouse [6] method is more accurate but even more time-consuming. In the GP method, the radiologist must look at the atlas images and estimate the bone age according to the similarities in the key points identified with the atlas image. This method is erroneous for children older than 14 years due to the high similarity of radiographic images of 14-19 years old [7]. The more complex White House Tanner method, in which the radiologist must score points for each key area individually and according to the atlas, and finally evaluate the scores to reach bone age. The advantages of the GP method are higher speed and the advantage of the Tanner White House (TW) method are more accuracy. The approximate time required to evaluate each method is 1.5 minutes and 8 minutes, respectively [8]. The evaluation error is 11.5 months and 9 months, respectively [9]. The GP method is much more widely used, for example in a study among members of the Society for Pediatric Radiology in the United States, 97% of them used the GP method for ages 3 to 18 using radiography of the wrist and hand [10].

Due to the time-consuming nature of these methods, and the possibility of human error in the assessment process, as well as the need for an experienced radiologist to achieve proper accuracy, makes a computer solution that performs the assessment automatically with appropriate speed and accuracy, is very practical. The ultimate goal is to use optimal and appropriate machine learning methods to increase the speed and accuracy of bone age prediction from left-hand radiographs. The result of this study can identify and prevent disorders in children and easier access to bone age. another thing that we tried to keep in mind was keeping everything small enough so that it could be run on normal computers that exist in medical places also the computer we tried to train models on was not powerful.

### 1.1. Dataset

In 2017, the North American Society of Radiologists released a collection of 14,236 left-hand radiographs of children to increase attention to bone age estimation, which varies in quality, size, and detail [11]. This data set consists of images, mainly from the left hand of children aged 1 month to 19 years, which is divided into three categories:

training dataset with 12611 images, development validation dataset with 1425 images, and test set with 200 images. For each radiograph, in addition to the exact age of the bone, the sex of the person is also stated, and due to the relationship between bone growth and structure to sex hormones, its use in the input data of the bone age estimation model is essential. The ratio of males-female in the Train dataset is 0.54 to 0.46 (5778-6833). As you can see, the frequency of images for different ages and genders is not equal, and this can create many challenges to achieve more accurate predictions.

### 1.2. Ideas for improving preprocessing

Despite the limited radiographic images for many age groups, the input images of the problem are not in their own right, depending on the devices and individuals, they have different characteristics such as accuracy, size, and considerable errors. To prevent the adverse effect of inappropriate images on the input, we need to build special tools to detect, improve accuracy and eliminate errors. Radiographic images can be imaged in two ways, which can have a significant impact on the accuracy of the final model due to the limited number of images in each age group. Input images are recorded at different angles. Given that the final CNN neural network model is sensitive to the rotation of input data, we need a tool that intelligently determines the angle of the hand in the image relative to the vertical side.

Due to the limited input size in deep neural networks, the efficient use of the input was very important for us. On the other hand, many images in RSNA dataset are taken at different distances and the hands in the radiographic images are not in standard condition. Due to the low contrast and complexity of the images, we have to improve the condition and quality of the input images by accurately detecting the location of the hand inside the input images.

## 2. RELATED WORKS

Bone age estimation methods can be divided into three parts; traditional methods are often performed by eye comparison. Image processing-based methods attempt to speed up the traditional method or to automatically estimate bone age by extracting basic features. Deep learning-based methods do the job of estimating bone age automatically by extracting more advanced and hard-to-detect features and relations.

### 2.1. Traditional methods

In the past, many attempts have been made to estimate bone maturity. The traditional methods introduced in the introduction, including the GP method and the TW method, are mostly based on atlases and are calculated by comparing images. But there are other methods, for example, Sauvegrain et al. [12] have introduced a method for determining bone age using radiographic images of the elbow, but here we focus on methods related to the wrist and fingers.

### 2.2. Image processing methods

Image processing-based methods have been used to speed up traditional methods or to extract features that are visually difficult. For example, E. Pietka et al. [13] obtains more information than visual comparison by identifying and extracting the epiphyseal/metaphyseal ROIs features. Gretych et al. [14], created a digital atlas of 1,400, which they used to estimate bone age by closely comparing incoming images with images in the atlas. F. Cao et al.[15], tried to solve the problems of the GP method with a software approach. By creating an atlas and a web-based service, they made it possible to estimate the age of the bone by comparing the input images and the images in the atlas. Thodberg et al. [16], introduced BoneExpert, a completely automatic solution was created without the need for a skilled radiologist. In this approach, using image processing and modeling the size and shapes of bones, radiographic images are divided into several parts, and the age of the bone is estimated. The method of calculating bone age makes the results convertible by GP and TW methods.

### 2.3. Deep neural network-based methods

Estimation of bone age with deep neural networks has been studied in various papers and researches and good information can be obtained from useful works and mistakes. Zhang et al. [17] , proposed the idea of integrating visual features with textual features to increase bone age estimation accuracy. The textual features are obtained from the results of radiologist analysis of 10 regions of interest, similar to the Tanner-Whitehouse method. However, since the RSNA dataset does not have those radiologist evaluations for each region of interest, only gender has been used in the textual data section for comparison with our research. The idea of integrating heterogeneous features is used in different parts of our method due to the significant impact it can have[18]. We use the integration of image features obtained from the CNN model and gender textual data. Also, the integration of layers with different sizes of the CNN model, as well as features obtained from different parts of the hand, greatly helps to increase the final accuracy. In the preprocessing process of this paper, simple methods have been used to identify the object, which leads to reduce their final accuracy due to the complexity of the images in the RSNA database.

Reddy et al. [19] compare the features of the index finger with the features of the whole hand. In this process, using the object recognition model. They separated the index finger and trained the model separately. They used RetinaNet [20] to detect index fingers but their final object detection model did not perform well and they crop 6% of the RSNA datasets images manually. Their final models were based on Xcecption [21] which uses depth-wise convolution layers to reduce the number of parameters and increase the model depth and both of them have been precisely competitive (the model that used the whole hand as input and the model that used only index finger). Based on the actions and results obtained from this paper, we concluded that we can use smaller parts of the radiographic image, such as the wrist and index finger, separately in the bone age estimation model. Of course, how to combine the results obtained from different parts of the radiographic image can be complicated. Therefore, using research related to the importance of each area of the wrist and fingers in estimating bone age can be helpful.

T.-Y. Lin et al. [22] examine the important areas that the bone age estimation model refers to at different ages. This work has been determined by examining the saliency map [23] and counting the ridge points in the five areas. Finally, it has been concluded that with age, the representative features of the bone age in the hand area will change. Based on these results, we decided to start our research by dividing each complete hand radiograph into similar parts. After processing the sections completely separately, the final model can more accurately estimate bone age according to the age ranges and features extracted by the smaller neural network models.

M. Escobar et al. [24] used an object recognition model to determine the position of the hand. In this method, the key areas related to the position of the hand are identified and used with a special pattern. By infusion visual radiographic data and hand position data the accuracy of the final model has increased. It should be noted that in the training process of the hand position recognition model, another data set has been used, which can be effective. The results of this study show that reducing the effect of hand position in the radiographic image can increase the effectiveness of extracting the relationship between bone age and features extracted by the bone age estimation model. Due to this result, in the following work, we try to make the radiographic images in the preprocessing stage similar in terms of their position in the image and their location. Also, in data augmentation, we reduce the effect of hand position by changing the perspective and warp part of the image randomly (like changing the shape or location of a finger).

L. Su et al. [25] focused on the preprocessing part, by using the Deeplab[26] model, they removed unwanted parts of the background, by doing this, they significantly increased the accuracy and quality of radiographic images and highlights bone age-related features. They turn the regression problem into a classification problem by dividing age into 3-month intervals. In the preprocessing stage, instead of usual data augmentation, they trained a GAN model to create new radiographic images. Of course, this is a difficult approach. Making artificial images is difficult, considering that the age of the newly created image must be the same as the age of the original image and there are not enough images for each age interval. This was a new method, but the final accuracy was not greater than the previous research. In this paper, the importance of data augmentation is examined, according to these results, we will create new artificial images more intelligently. Also, in preprocessing using the Deeplab model, we can remove unwanted parts of radiographic images. By removing unwanted parts, operations such as histogram equalization can improve the quality of images and make them more similar in terms of brightness. Since brightness is not related to bone age, the similarity of radiographic images in terms of brightness can help to find bone-related features more quickly.

Based on previous papers, we tried to use their strengths and not repeat the weaknesses. For this reason, in the preprocessing section, we optimally remove the background, optimally perform the data augmentation process, and finally improve the results with the division and solution approach.

## 3. METHOD

Based on related works, it can be concluded that the preprocessing process has an undeniable effect on the accuracy of the final model. On the other hand, in the process of preprocessing and preparing the input data, the occurrence of errors can cause the loss of necessary information and features and ultimately have a direct impact on the accuracy of the bone age estimation model. We carefully analyzed each preprocessing step to prevent errors from accumulating.

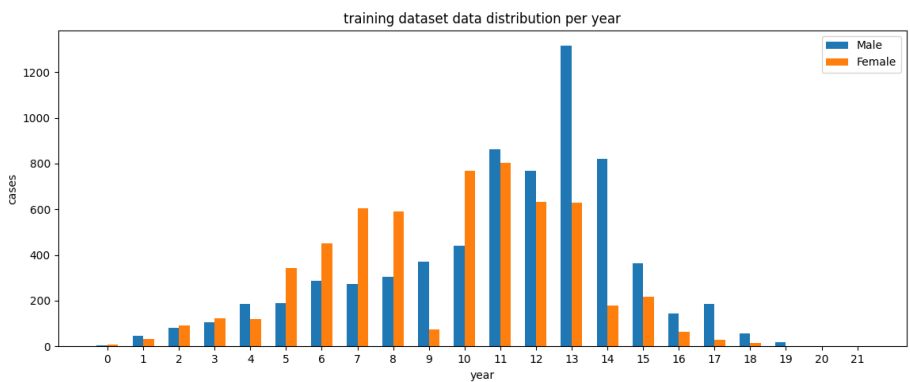

Fig. 1. In the above two graphs, we can see the unbalanced distribution of data, for some age groups, there is little or no data. In some age groups, the gender balance is very high and can affect the final results.

According to Figure 1, we can easily detect the unbalanced distribution of data set data. There are few samples for ages under 12 months and over 216 months (18 years). We know that in these two categories, diagnosing bone age is less important. After boys reach the age of 16 and girls reach the age of 15, 99 percent of their longitudinal bone growth has taken place [27]. At the age of under 12 months old, there is little information about the shape of the bones [12]. Also, in the 200 images of the test set, there is only one image related to each of these two categories. Therefore, in the following research, we will focus on the traditional range of 1–18-year-old cases.

Examination of the number of samples of radiographic images for each age group shows that there is a big difference between different ages. In the preprocessing process, we try to achieve balance distribution in all age ranges by making more artificial images for the categories that have few samples. We did this with a small amount of rotation, zoom, transformation, grid distortion, and changing perspective.

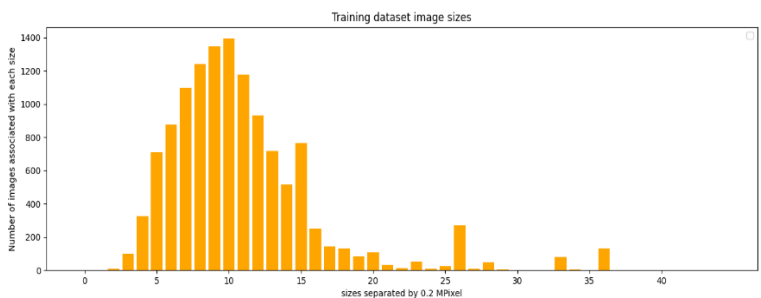

Fig. 2. The diagram shows the extent of the difference in the size of the RSNA dataset images.

Due to the variety and quality and structures of RSNA radiographic images, visible in Figure 2, object detection models are not sufficient to separate desired parts. in many images, after histogram equalization, the noise will destroy the main features, and binarization with thresholding doesn't work well. For this purpose, we need to train the neural network model that can separate the desired areas from the unwanted parts with very high accuracy. Here we use the DeeplabV3Plus model [28]. To train the Deeplab model and remove background noise from images, we need a training dataset of radiographic images. Due to the variety of images in the RSNA database, it is essential to build a new dataset from highly detailed masks. So, we selected 751 images and made very high-precision masks for each one of them with the help of image processing software. The masks we made are the expected output of the Deeplab model, some examples of these images and their associated masks are shown in Figure 3. In the next step, by applying data augmentation simultaneously on the created masks and respected images, we created a suitable dataset for training the Deeplab model. After training the results on images that the model had not seen before were extremely accurate.

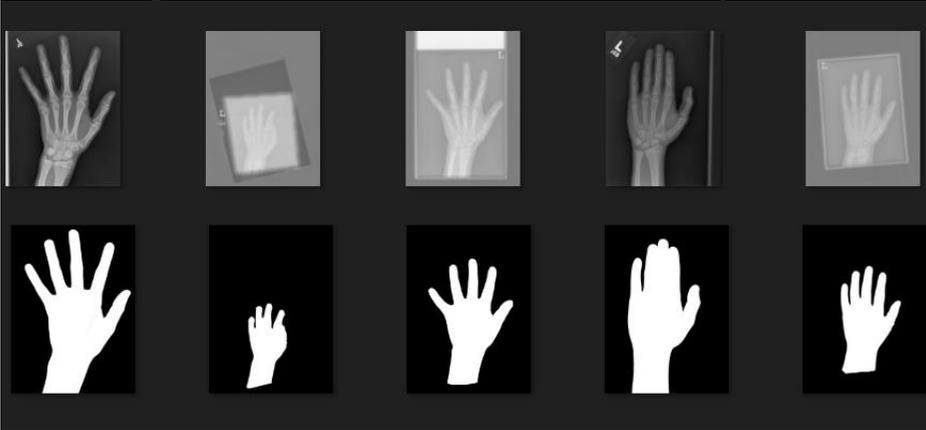

Fig. 3. The first row shows examples of RSNA dataset images and the second row shows the output of each image from Deeplab model after direction and angle being fixed.

Convolutional neural networks, with filter design, have a special ability to extract image features regardless of the relative displacement of the main subject, however, the sensitivity of CNN neural networks to image rotation is a known problem[29], and various methods have been proposed in the past to address this issue[30]. Due to the presence of images from different angles in the RSNA dataset and the lack of value in detecting bone age from different angles, we attempted to build a separate angle detection model to correct the angle of the images. Of course, a small difference in the angle of the images does not cause a problem, but here we are talking about 30 to 180 degrees. Similarly, due to the possibility of imaging radiographic results from two directions, we created another model to detect the direction of the images.

The two models we built are very simple and based on MobileNetV2. Our main focus in training these two models was the data augmentation process. Given that we defined each model for a specific task, to train them, we were able to focus on the main task of each model and perform data augmentation depending on the task. For this purpose, we selected 751 images from the RSNA training dataset and gave each one an angle label and an orientation label. In the next step, for data augmentation of the angle detection model, we focused on changing the angles with the label to cover all the angles. To train the direction detection model, in addition to other methods, we flipped all the images and the label with respect to each other. Given that the trained models had only one task, their results were obtained with great accuracy. The result of this process will reduce the complexity of the object detection model inputs and age estimation model inputs.

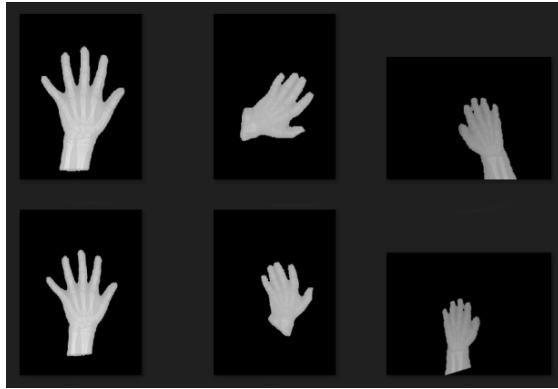

Fig. 4. The first row shows the raw output of RSNA images after being masked with Deeplab model output, and the second row shows repaired images by angle detection model output and direction detection model outputs.

Instead of using a large model with a big input size to estimate bone age, we divide the problem into smaller sections using the divide and conquer strategy. Initially, according to Figure 5-a, we selected five areas for further processing, the specified areas should be identified by an object recognition model. We created a model based on EfficientDet [31] architecture, for the object recognition task, after creating a new dataset using the detected regions, we will use smaller regions to predict bone age. We used the EfficientNetB0[32] base model for the object recognition model, as well as three custom BiFPN layers similar to what shown in Figure 7, and finally several outputs for specifying each region location.

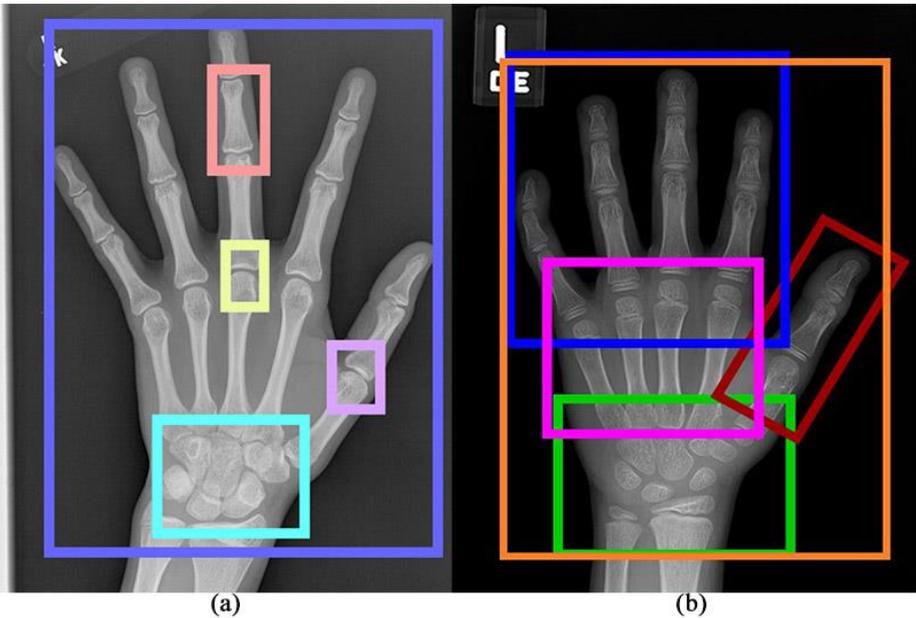

Fig. 5. The marked color areas must be identified separately by the object recognition model. (a) Smaller areas are used in the final solution. (b) Areas selected to start research according to related work.

The method we present is shown in summary in Figure 6. First, using the Deeplab model, we removed the extra areas of the image and by removing unnecessary parts of images, we could improve the light contrast and image clarity. Then we correct the angle and direction of the image, and by using an object detection model we divided the main image into smaller and more important parts, and finally, by using a regression model, we try to find the relationship between the bone visual features from input images and the age of the bone.

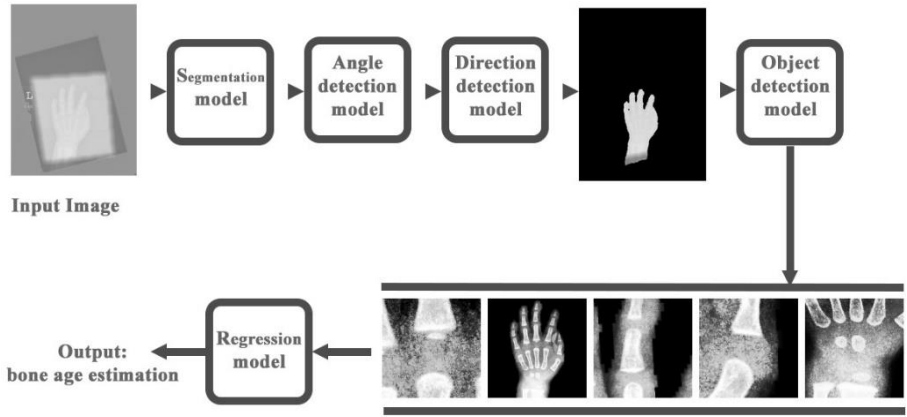

Fig. 6. Shows the steps for estimating bone age in the proposed method. In this method, we focus on the smaller parts of the input image, and use the large image to preserve information related to hand posture.

## 4. EXPERIMENTS

According to related work, we knew that different parts of the radiographic image are of different value and importance at different ages. That's why we started with areas shown in Figure 5-b, to begin with. According to the size of the images in the RSNA data set that can be seen in Figure 2 and different distances from the radiographic images that can be seen in the first row of Figure 3, 224 x 224 input sizes were selected for each region. By training a model with the structure shown in Figure 7, we crated the regression model presented in Figure 7 and estimated bone age.

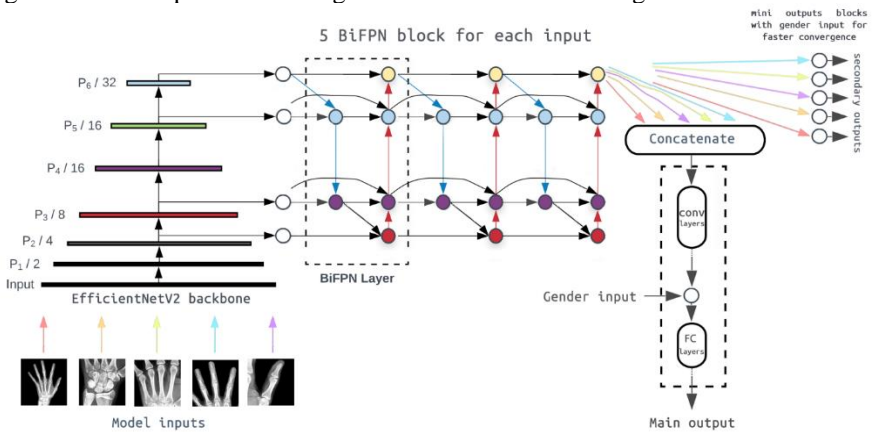

Fig. 7. The figure shows the structure of the initial model. Colored arrows indicate separate paths from input to output.

In the initial model shown in figure 7, we used the pre-trained EfficientNetV2B0 [33] model on the imagenet [34] dataset as a backbone. This model, despite being less complex, has the same performance as Inception-V3 [35] and is compatible with the input size we chose. Then we used several layers of BiFPN, these layers work very well for transfer learning. In addition to the main model output, we created a separate output for each input, using these outputs for further analysis and faster convergence of the model. After training the model, the best MAE error reached was 4.29 months so by using Saliency Maps [23], we started evaluating the images and their important regions.

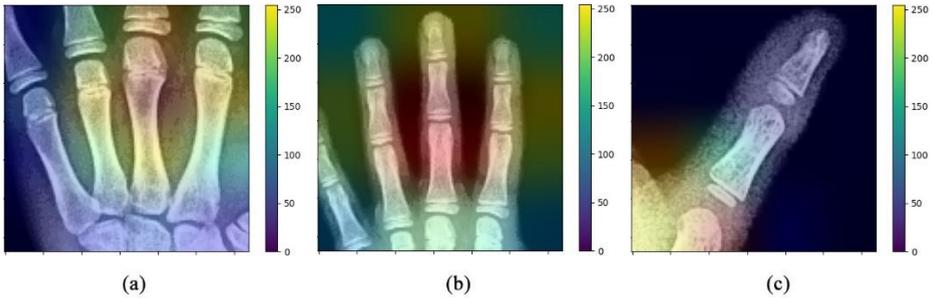

Fig. 8. By using grad-cam, we examined the important areas of the input, when we got good results from our bone age assessment model. the blue areas are less important and the red and yellow areas are more important. (a) The model often pays attention to the middle finger for accurate estimation, which may be because in preprocessing the angle of the images is aligned with the middle finger, resulting in better features being extracted from this area. The model paid less attention to the upper part of the wrist. (b) The model pays more attention to the middle finger than other parts. Attention decreases for the lower part of the fingers and the upper part of the fingers and increases in the middle part. (c) The model did not pay attention to the upper parts of the thumb, but by paying attention to the lower part of the thumb accurate results were estimated.

by using Grad-Cam[36] we can determine, by looking at which part of the input image, the bone age estimation model predicted the final output. Therefore, after examining the cases where the model was very accurate, we identified new areas in Figure 5a. There was a big difference between the initial areas and the important areas, some examples of which are shown in Figure 8. In Figure 8b, other parts of the fingers can also be considered, but the middle finger is well-photographed and is set vertically by the rotation detection model in most of the RSNA datasets images. But for other areas such as the wrist and the whole hand, there was not much difference, so we did not make much change in these regions.

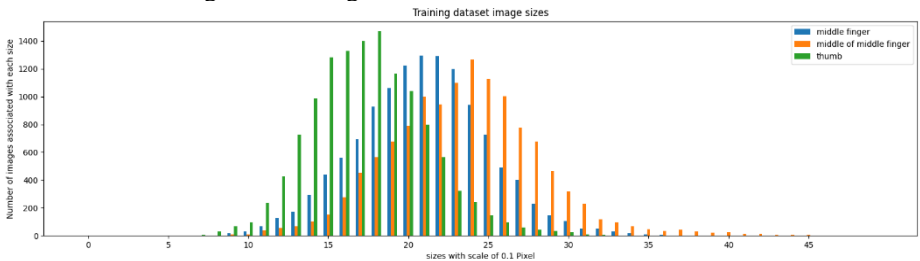

Fig. 9. The image size diagram shows the input images cropped by new object recognition model outputs. Note that for data augmentation, the sizes are little larger than the final size and the width and height of the images are equal.

After selecting the new regions, we created a data set for each region using 751 images and trained the object recognition model to identify and cut new regions. After shrinking the selected regions, due to the small size of many RSNA dataset images according to Figure 9, the size of many regions was smaller than 224 x 224 pixels, but most new deep neural network models do not get good results with small input sizes. And upscaling images only creates a larger blurred image. So, we changed the input size to 128 x 128, which reduced the model's complexity and allowed us to do a lot more data augmentation. To prevent the model from being overfitted in the training process in each epoch, we randomly selected a part of the training data set. But no progress was made using the same structure as before.

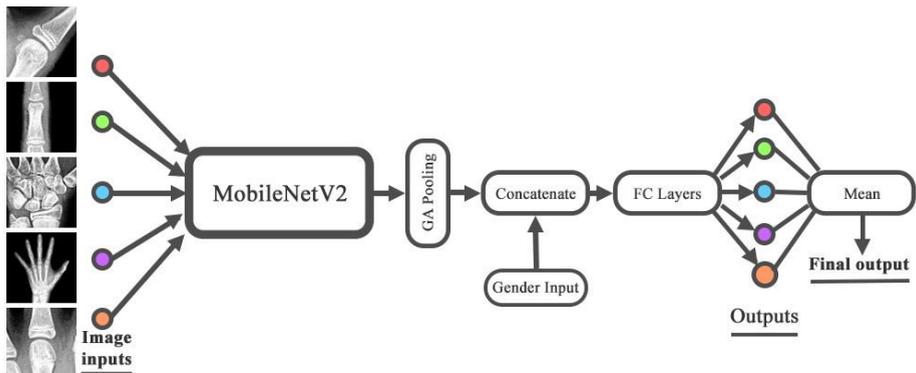

Fig. 10. Shows the structure of the final model, each input is processed separately. In the final step, by calculating the mean of 5 outputs, the final output is obtained.

Consequently, we looked at different structures, and finally, we tried MobileNetV3[37] but pre-trained weights with 128x128 pixels were not available and the model did not reduce the error on the RSNA test set compared to other methods. we created a model based on MobileNetV2 [38] that received all the regions as input. This neural network model performed very well with 128 x 128 input size and its pre-trained weights were available. Due to the small size of the model, it was possible to stack several of them. Figure 10 shows the structure of the final model. In this structure, we use the images of different regions separately as input, after extracting features with MobileNetV2 we use global average pooling[39] and concatenate extracted features with gender information. Then its fully connected layers job to extract the effect of gender on visual features and convert them to a single number as bone age. Instead of one or more weighty layers, we used the mean in the final step to prevent the model from overfitting the training data. The results obtained by this model were very good and the errors of some inputs were corrected by the final stage mean layer. We started training with pre-trained Tensorflow [40] weights based on the Imagenet dataset for the MobileNetV2 base model that we used, we froze the weights of the base model, trained other layers of the model, and then made the whole model trainable. We used Adam [41] as an optimizer and Mean Square Error as a loss function and a 1e-4 learning rate to train the model. All experiments are implemented on a computer with specifications: Intel® Core™ i7-4790 CPU, 16 GB RAM, Nvidia GeForce GTX 960 GPU.

## 5. RESULT AND ANALYSIS

To show the progress and effects of the actions we have taken, we compare our introduced models and the results of related works. In the process of research, we found two suitable models, the first model that we used for analyzing is beaten up by the second model in terms of performance, but compared to most related work results, it is still competitive, we named the first model EfficientNetV2B0 based model and the second model MobileNetV2 based model. Table 1 shows a comparison of these two methods based on the RSNA test dataset. In the stacked version, we trained the MobileNetV2-based model three times separately, averaged their predictions, and used it as a final estimate.

Table.1. Comparison between EfficientNetV2B0 based model and MobileNetV2 based model, on the RSNA test set.

| Model | MAE Error | Range (year) |
|---|---|---|
| EfficientNetV2B0 based model | 4.29 | 0-20 |
| MobileNetV2 based model | 3.97 | 0-20 |
| MobileNetV2 based model | 3.90 | 1-18 |
| MobileNetV2 based model stacked | 3.90 | 0-20 |
| MobileNetV2 based model stacked | **3.84** | 1-18 |

In the training process, we have concentrated on the range of 1-18 years and the estimate of the age of the bone outside this age group is not very precise. We have still obtained acceptable results, and there is not much difference between the error obtained in the period from 1 to 18 years and the period from 0 to 20 years. Table 2 provides a summary of the comparison of the results achieved and the related work results.

Table.2. Comparison of the results obtained in this study and related work on RSNA test set.

| Model | MAE Error |
|---|---|
| F. Chollet[22] | 8.66 |
| P.Hao [17] | 6.2 |
| N. E. Reddy[19] | 4.7 |
| RSNA Challenge winner | 4.26 |
| CNN-GAN-OTD[25] | 4.23 |
| BoNet[24] | 4.14 |
| **Our MobileNetV2 based model** | **3.90** |

After seeing the comparison results in Table No. 2, we can conclude that it is enough to extract features from the areas introduced by us to accurately estimate bone age. Of course, our pre-estimation actions had a great impact in reaching this result, for example, removing the background of the images with the help of the model we created for this task greatly helped to increase the contrast and details of the images. Also, using the angle detection model made the input images more similar and removed unwanted complexity. The presented object detection model was modeled after the EfficientDet model, but it is much simpler than the original model, however, it provides very good performance and can be used in related research when the desired objects are definitely present in the input.

According to Table 2, it can be concluded that the final model has acceptable performance, although the error obtained is not the same in the entire range of 0 to 19 years. For example, we examine the MobileNetV2 based model that has an error of 3.90 months in the period of 1-18 years on the RSNA test set, in Figure 11. According to the Mean Error, it can be seen that in the period of 10 to 18 years, the annual error is often less than 3.90 months. It should be noted that in the preprocessing process, we tried to reduce the error of this interval by performing more data augmentation over the age range of 1 to 10 years, however, the error did not fall below this limit. Due to the rapid growth of children in the range of 3 to 10 years and according to Figure 1, the smaller number of samples, more data is needed to reduce errors in this age range.

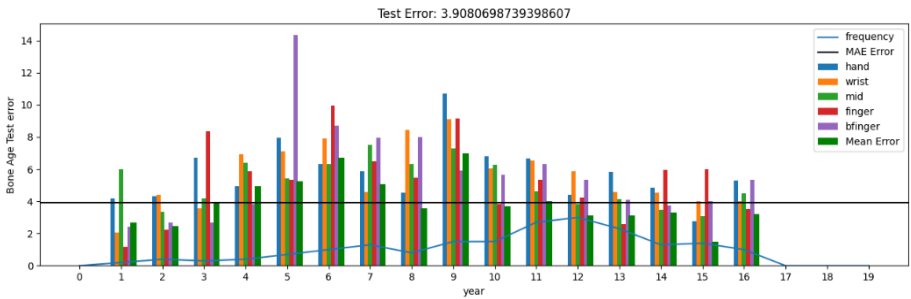

Fig. 11. The bar chart shows the errors obtained at the outputs and the final output of the non-stacked MobileNetV2-based bone age estimation model. According to the frequency diagram, we can see the frequency of samples and the importance of the error associated with each year.

Another interesting case is the observation of larger output errors for each region than the average error in Figure 11, it can be noted in many cases, all outputs reach a low error of estimation, however, at some ages, some regions are similar to older age regions and some are similar to younger age regions. For example, Figure 13 shows a case with very low estimation error between the actual value and the mean output value, but the predicted age is larger with respect to the whole hand image and smaller with respect to the middle finger proximal phalanx and wrist. Similar cases also exist with this feature, which show the correlation of the regions selected by us to achieve high accuracy.

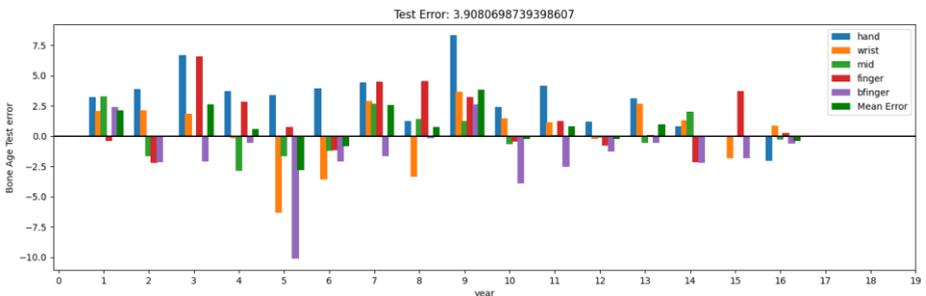

Fig. 12. The graph shows the average error of each age category in the RSNA test set. The bars that are below the line mean that on average the predicted value was lower than the actual value.

In related research, it was pointed out that at different ages there is a clear bias to estimate a higher or lower bone age from different parts of the hand. According to Figure 12, it can be concluded that the big toe is usually predicted to be less than the actual value, this difference decreases as children grow older. In our method, the image of the hand was used to consider the general state of the hand as well as the position of the hand. Due to the size of the input images and the lack of details of small bones, at younger ages, a larger value for bone age is usually estimated using the general image of the hand, although, at older ages, this difference is less.

Regarding the middle finger, especially in the age range of 1 to 9 years, the estimated age is usually greater than the actual age, but after that, it is estimated with high accuracy. The wrist has different biases in the estimation of bone age in different age ranges, but on average, according to the radiographic images of the wrist before the age of 9, lower values and then higher values are estimated.

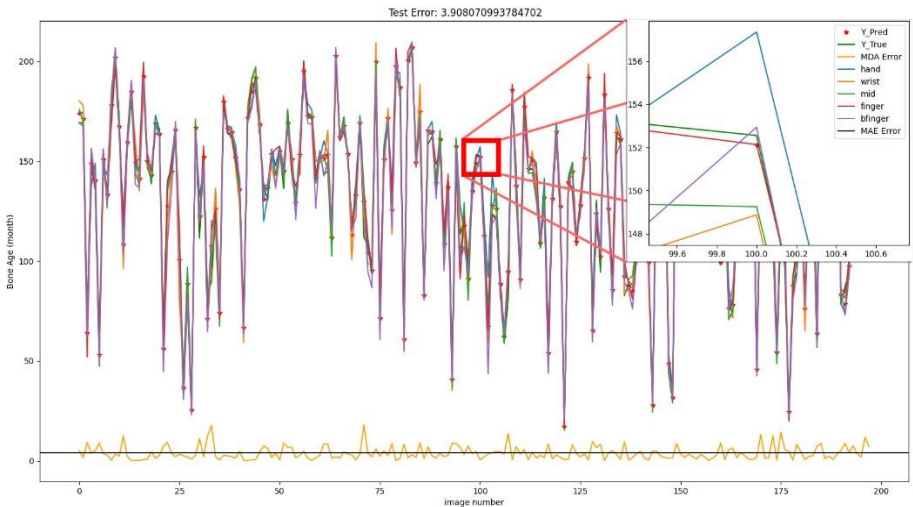

Fig. 13. The diagram shows the output predictions of the non-stacked MobileNetV2-based bone age estimation model on the RSNA test set. The red star indicates the actual value and the lines shows the predicted values.

## 6. DISCUSSION

Based on the results, it can be concluded that by using the key points introduced by us, it is possible to estimate bone age faster than traditional methods such as TW2 or GP. Although the details may be difficult for radiologists to discern. The reduction of the number of key areas creates the limitation that if the areas we have chosen are incomplete in the radiographic image, it is possible that the final model cannot estimate bone age with high accuracy. If the state of the input image is not suitable and the object recognition model faces problems, probably the skilled radiologist will also face problems in dealing with the problematic radiographic image.

Bone age estimates are often used to predict children's final height, and parents are usually more concerned about final height after the age of 10, so there are more samples available for ages older than 10 years [42]. In the age range older than 10 years, the model we built has a very low error. In our evaluations, a large number of images related to ages younger than 10 years had less quality and detail. In future research, if we can create an upscaling model with the ability to predict bone tissues, we can more accurately estimate bone age in such radiographic images.

## 7. CONCLUSION

In this study, we assessed the bone age estimation field, first we evaluated the basic methods of estimating bone age. Then evaluated the approaches used in previous research, and then based on the results of previous papers, new methods of object detection, and improvement of the preprocessing, we created an initial model. By evaluating the initial model and analyzing the points that the model paid attention to, we introduced suitable areas for bone age estimation, the number of these areas is less than previous methods such as TW and GP. By using the new areas, we were able to present a model with much higher accuracy than the previously introduced models, even though the model presented by us is much lighter in terms of processing and can be used in the backend of web services to estimate bone age. In the end, we reached an MAE of 3.90 months in the range of 0-20 years and an MAE of 3.84 months in the range of 1-18 years.

## 8. DECLARATIONS

No funds, grants, or other support was received.